\documentclass[11pt,a4paper]{article}
\usepackage{times,latexsym}
\usepackage{url}
\usepackage[T1]{fontenc}
\usepackage{amsmath}
\usepackage{amssymb}
\usepackage{tabularx}
\usepackage{mathtools}
\usepackage{booktabs}
\usepackage{url}
\usepackage{longtable}
\usepackage{tabu}
\usepackage{multirow}
\usepackage{amsfonts}
\usepackage{tabu}
\usepackage{algorithm}
\usepackage{bbm}
\usepackage{subfigure}
\usepackage[noend]{algpseudocode}
\usepackage[normalem]{ulem}
\usepackage{enumitem}
\makeatletter

\def\BState{\State\hskip-\ALG@thistlm}
\usepackage{bbm}
\usepackage{xcolor}
\DeclareMathOperator*{\argmax}{arg\,max}

\DeclareMathOperator*{\argmin}{arg\,min}

\usepackage[acceptedWithA]{tacl2018v2}
\usepackage{xspace,mfirstuc,tabulary}

\newif\iftaclinstructions
\taclinstructionsfalse 
\iftaclinstructions
\renewcommand{\confidential}{}
\renewcommand{\anonsubtext}{(No author info supplied here, for consistency with
TACL-submission anonymization requirements)}
\newcommand{\instr}
\fi

\iftaclpubformat 

\else

\fi

\title{Reducing Confusion in Active Learning for Part-Of-Speech Tagging}

\author{Aditi Chaudhary\textsuperscript{1}, 
    Antonios Anastasopoulos\textsuperscript{2,\Thanks{ Work done at Carnegie Mellon University.}},
     Zaid Sheikh\textsuperscript{1}, Graham Neubig\textsuperscript{1} \\
  \textsuperscript{1}Language Technologies Institute, Carnegie Mellon University\\
  \textsuperscript{2}Department of Computer Science, George Mason University\\
  { \texttt{\{aschaudh,zsheikh,gneubig\}@cs.cmu.edu}} \hspace{.5cm}
  { \texttt{antonis@gmu.edu}}
 }

\date{}

\begin{document}
\maketitle
\begin{abstract}
Active learning (AL) uses a data selection algorithm to select useful training samples to minimize annotation cost. This is now an essential tool for building low-resource syntactic analyzers such as part-of-speech (POS) taggers. Existing AL heuristics are generally designed on the principle of selecting \emph{uncertain} yet \emph{representative} training instances, where annotating these instances may reduce a large number of errors. However, in an empirical study across six typologically diverse languages (German, Swedish, Galician, North Sami, Persian, and Ukrainian), we found the surprising result that even in an \emph{oracle} scenario where we know the true uncertainty of predictions, these current heuristics are far from optimal. Based on this analysis, we pose the problem of AL as selecting instances which \emph{maximally reduce the confusion between particular pairs of output tags}.
Extensive experimentation on the aforementioned languages shows that our proposed AL strategy outperforms other AL strategies by a significant margin. 
We also present auxiliary results demonstrating the importance of proper calibration of models, which we ensure through cross-view training, and analysis demonstrating how our proposed strategy selects examples that more closely follow the oracle data distribution. The  code is publicly released here.\footnote{\url{https://github.com/Aditi138/CRAL}}
 
\end{abstract}

\section{Introduction}

Part-Of-Speech (POS) tagging is a crucial step for language understanding, both being used in automatic language understanding applications such as named entity recognition (NER; \citet{nazeer2018part}) and question answering (QA; \citet{wang2018structured}), but also being used in \emph{manual} language understanding by linguists who are attempting to answer linguistic questions or document less-resourced languages \cite{anastasopoulos2018part}.   Much prior work \cite{huang2015bidirectional,bohnet2018morphosyntactic} on developing high-quality POS taggers uses neural network methods which rely on the availability of large amounts of labelled data. However, such resources are not readily available for the majority of the world's 7000 languages \cite{hammarstrom2018glottolog}.  Furthermore, manually annotating large amounts of text with trained experts is an expensive and time-consuming task, even more so when linguists/annotators might not be native speakers of the language. 
\begin{figure*}[t]
\small
\centering
\includegraphics[width=\textwidth]{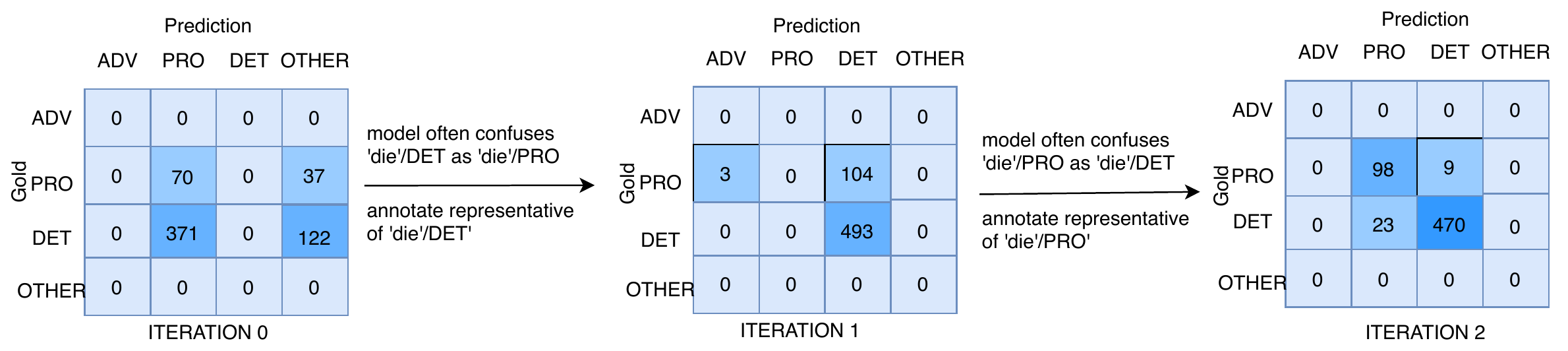}
\vspace{-1em}
\caption{\label{eg} Illustration of selecting representative token-tag combinations to reduce confusion between the output tags on the German token `die' in an idealized scenario where we know true model confusion.}
\vspace{-1em}
\end{figure*}

Active Learning \cite[AL]{lewis1995evaluating,settles2009active} is a family of methods that aim to train effective models with less human effort and cost by selecting such a subset of data that maximizes the end model performance. While many methods have been proposed for AL in sequence labeling \cite{settles2008analysis, marcheggiani2014experimental, fang2017model}, through an empirical study across six typologically diverse languages we show that within the same task setup these methods perform inconsistently. Furthermore, even in an \emph{oracle} scenario 
where we have access to the true labels during data selection, existing methods are far from optimal.

We posit that the primary reason for this inconsistent performance is that while existing methods consider uncertainty in predictions, they do not consider the \emph{direction} of the uncertainty with respect to the output labels. For instance, in Figure \ref{eg} we consider the German token ``die,'' which may be either a pronoun (PRO) or determiner (DET).
According to the initial model (iteration~0), ``die'' was labeled as PRO majority of the time, but a significant amount of probability mass was also assigned to other output tags (OTHER) for many examples.
Based on this, existing AL algorithms that select uncertain tokens will likely select ``die'' because it is frequent and its predictions are not certain, but they may select an instance of ``die'' with \emph{either} a gold label of PRO or DET.
Intuitively, because we would like to correct errors where tokens with true labels of DET are mis-labeled by the model as PRO, asking the human annotator to tag an instance with a true label of PRO, even if it is uncertain, is not likely to be of much benefit.

Inspired by this observation, we pose the problem of AL for part-of-speech tagging as selecting tokens which maximally \emph{reduce the confusion} between the output tags.
For instance, in the example we would attempt to pick a token-tag pair ``die/DET'' to reduce potential errors of the model over-predicting PRO despite its belief that DET is also a plausible option.
We demonstrate the features of this model in an oracle setting where we know true model confusions (as in Figure \ref{eg}), and also describe how we can approximate this strategy when we do not know the true confusions.

We evaluate our proposed AL method by running simulation experiments on six typologically diverse languages namely German, Swedish, Galician, North Sami, Persian, and Ukrainian, improving upon models seeded with cross-lingual transfer from related languages \cite{cotterell-heigold-2017-cross}. In addition, we conduct human annotation experiments on Griko, an endangered language that truly lacks significant resources.  
Our contributions are as follows:
\begin{enumerate}[leftmargin=*,nolistsep,noitemsep]
    \item We empirically demonstrate the shortcomings of existing AL methods under both conventional and ``oracle'' settings. Based on the subsequent analysis, we propose a new AL method which achieves +2.92 average per-token accuracy improvement over existing methods under conventional settings, and a +2.08 average per-token accuracy improvement under the \textit{oracle} setting.
    \item We conduct extensive analysis measuring how the selected data using our proposed AL method closely matches the oracle data distribution.
    \item We further demonstrate the importance of model calibration, the accuracy of the model's probability estimates themselves, and demonstrate that cross-view training \cite{clark2018semi} is an effective way to improve calibration.
    \item We perform human annotation using the proposed method on an endangered language, Griko, and find our proposed method to perform better than the existing methods. In this process, we collect 300 new token-level annotations which will help further Griko NLP. 
\end{enumerate}

\section{Background: Active Learning}
\label{sec:exwork}
Generally, Active Learning (AL) methods are designed to select data based on two criteria: ``informativeness'' and  ``representativeness'' \cite{huang2010active}. Informativeness represents the ability of the selected data to reduce the model uncertainty on its predictions, while representativeness measures how well the selected data represent the entire unlabeled data. AL is an iterative process and is typically implemented in a batched fashion for neural models \cite{sener2017active}. In a given iteration, a batch of data is selected using some heuristic on which the end model is trained until convergence. This trained model is then used to select the next batch for annotation, and so forth. 

In this work we focus on \emph{token-level} AL methods which require annotation of individual tokens in context, rather than full sequence annotation which is more time consuming. Given an unlabeled pool of sequences $D\!=\!\{\mathbf{x_1}, \mathbf{x_2}, \cdots, \mathbf{x_n}\}$ and a model~$\theta$, $P_{\theta}(y_{i,t}\!=\!j \mid \mathbf{x_i})$ 
denotes the output probability of the output tag $j \in \mathcal{J}$ produced by the model $\theta$ for the token $x_{i,t}$ in the input sequence $\mathbf{x_i}$. $\mathcal{J}$ denotes the set of POS tags.  Most popular methods \cite{settles2009active, fang2017model} define the ``informativeness'' using either \emph{uncertainty sampling} or \emph{query-by-committee}. We provide a brief review of these existing methods.
\begin{itemize}[leftmargin=*]
 \item \textbf{Uncertainty Sampling} (\textbf{\textsc{uns}}; \citet{fang2017model}) selects the most uncertain word types in the unlabeled corpus $D$ for annotation. First, they calculate the token entropy $H(x_{i,t};\theta)$ for each unlabeled sequence $\mathbf{x_i} \in D$ under model $\theta$, defined as
 \begin{align*}
 p_{i,t,j} := & P_{\theta}(y_{i,t} = j \mid \mathbf{x_i}) \\
 H(x_{i,t};\theta) = & - \sum_{j\in \mathcal{J}} p_{i,t,j} \log p_{i,t,j}
 \end{align*}
 Next, this entropy is aggregated over all token occurrences across $D$ to get an uncertainty score $S_\textsc{uns}(z)$ for each word type $z$:
 \begin{equation*}
     S_{\textsc{uns}}(z) = \sum_{\mathbf{x_i} \in D} \sum_{x_{i,t}=z} H(x_{i,t}; \theta)
 \end{equation*}
    \item \textbf{Query-by-commitee} (\textbf{\textsc{qbc}}; \citet{settles2008analysis}) selects the tokens having the highest disagreement between a committee of models $C = \{ \theta_{1}, \theta_{2}, \theta_{3}, \cdots\}$ which is aggregated over all token occurrences. The token level disagreement scores are defined as
    \begin{equation*}
    S_{\textsc{dis}}(x_{i,t}) = |C| - \max \!\sum_{y \in [\hat{y}_{i,t}^{\theta_{1}}, \hat{y}_{i,t}^{\theta_{2}}, \cdots, \hat{y}_{i,t}^{\theta_{c}}]}\!V(y),
    \end{equation*}
    where $V(y)$ is number of ``votes'' received for the token label $y$.  $\hat{y}_{i,t}^{\theta_{c}}$ is the prediction with the highest score according to model $\theta_{c}$ for the token $x_{i,t}$. 
    These disagreement scores are then aggregated over word types:
    \begin{equation*}
        S_{\textsc{qbc}}(z) = \sum_{\mathbf{x_i} \in D} \sum_{x_{i,t}=z} S_{\textsc{dis}}(x_{i,t})
    \end{equation*}
\end{itemize}
Finally, regardless of whether we use an uncertainty-based or QBC-based score, the top $b$ word types with the highest aggregated score are then selected as the to-label set
 \begin{equation*}
     X_{\textsc{label}} = {\rm b\text{-}}\argmax_z S(z),
 \end{equation*}
 where ${\rm b\text{-}}\argmax$ selects top $b$ word types having the highest $S(z)$.
\citet{fang2017model} and \citet{chaudhary-etal-2019-cmu} further attempt to include the ``representativeness'' criterion by combining uncertainty sampling with a bias towards high-frequency tokens/spans. 

\paragraph{Failings of current AL methods}
While these methods are widely used, in a preliminary empirical study we found that these existing methods are less-than optimal, and fail to bring consistent gains across multiple settings. Ideally, having a single strategy that performs the best across a diverse language set is useful for other researchers who plan to use AL for new languages. Instead of them experimenting with different strategies with human annotation, which is costly,  having a single strategy known a-priori will reduce both time and human annotation effort. Specifically, we demonstrate this problem of inconsistency through a set of \emph{oracle} experiments, where the data selection algorithm has access to the true labels.
These experiments hope to serve as an upper-bound for their non-oracle counterparts, so if existing methods do not achieve gains even in this case, they will certainly be even less promising when true labels are not available at data selection time, as is the case in standard AL.  

 \begin{figure}[t]
\small
\centering
\includegraphics[width=\columnwidth]{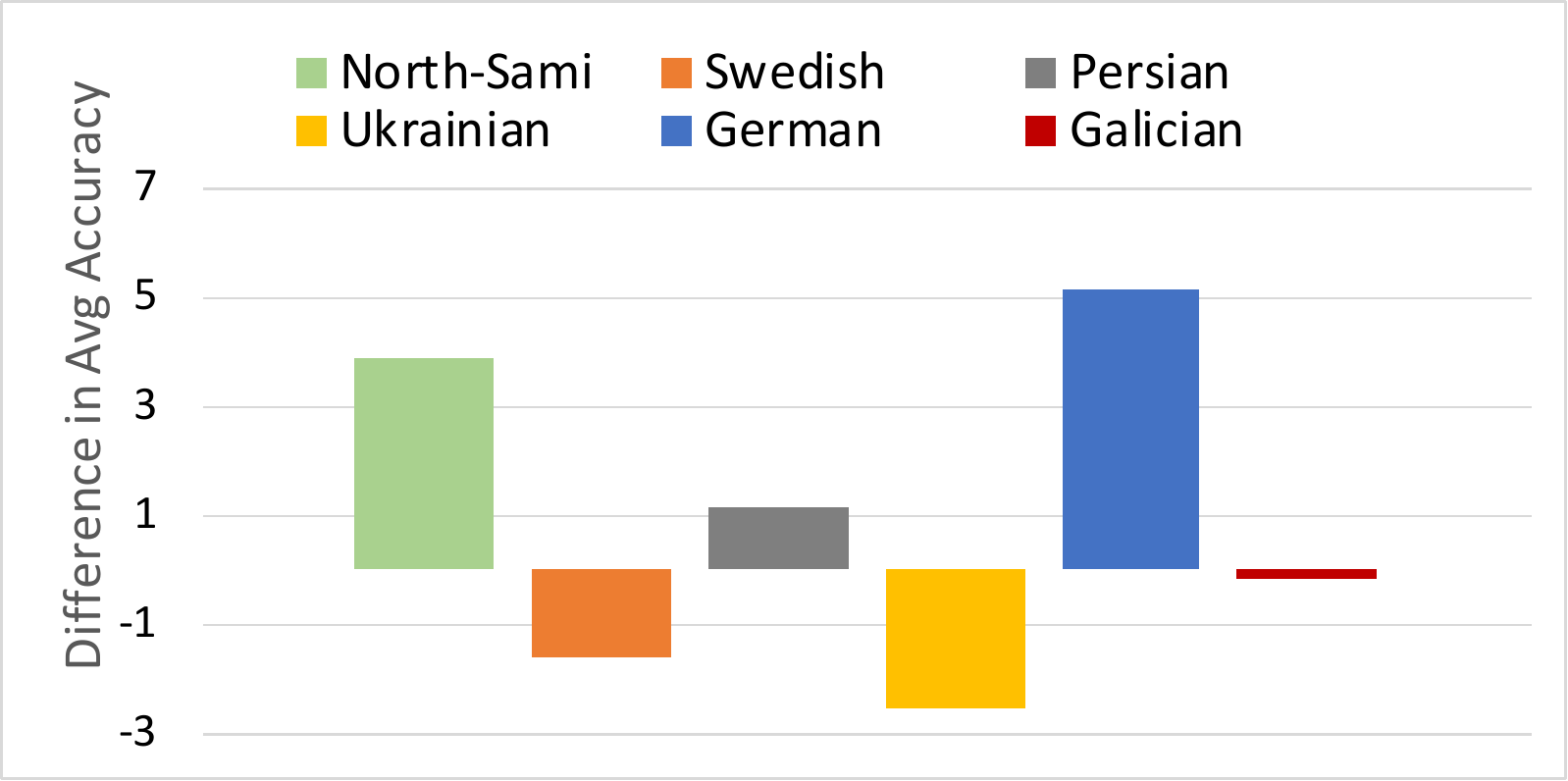}
\vspace{-1em}
\caption{\label{oracle}  Illustrating the inconsistent performance of \textsc{uns-oracle} and \textsc{qbc-oracle} methods. y-axis is difference in the POS accuracy for these two methods, averaged across 20 iterations having a batch size 50. }
\vspace{-1em}
\end{figure}

Concretely, as an oracle \emph{uncertainty sampling} method \textsc{uns-oracle}, we select types with the highest negative log likelihood of their true label.
As an  ``oracle'' \emph{query-by-committee} method \textsc{qbc-oracle}, we select types having the largest number of incorrect predictions.
We conduct 20 AL iterations for each of these methods across six typologically diverse languages.\footnote{More details on the experimental setup in Section~\S\ref{sec:setup}.}

First, we observe that between the oracle methods (Figure \ref{oracle}) no method consistently performs the best across all six languages. Second, we find that just considering uncertainty leads to unbalanced selection of the resulting tags. To drive this point across, Table~\ref{tab:iteration} shows the output tags selected for the German token `zu' across multiple iterations. \textsc{uns-oracle} selects the most frequent output tag, failing to select tokens from other output tags.  Whereas \textsc{qbc-oracle} selects tokens having multiple  tags, the distribution is not in proportion with the true tag distribution.  Our hypothesis is that this inconsistent performance occurs because none of the methods consider the confusion between output tags while selecting data. This is especially important for POS tagging because we find that the existing methods tend to select highly syncretic word types. \emph{Syncretism} is a linguistic phenomenon where distinctions required by syntax are not realized by morphology, meaning a word type can have multiple POS tags based on context.\footnote{Details can be found in Section \S\ref{sec:qual}, Table~\ref{tab:qual}.} 
 This is expected because syncretic word types, owing to their inherent ambiguity, cause high uncertainty which is the underlying criterion for most AL methods.

\begin{table}[t]
\centering
\resizebox{\columnwidth}{!}{
 \begin{tabular}{c|l|l}
  & \textsc{qbc-oracle}  & \textsc{uns-oracle} \\
 \midrule
\textbf{\textsc{iteration-1}} &  PART=1  &  ADP=1 \\
\textbf{\textsc{iteration-2}} & 	PART=1,ADP=1 & ADP=2  \\
\textbf{\textsc{iteration-3}} & ADV=1,PART=1,ADP=1 & ADP=2 \\
\textbf{\textsc{iteration-4}} &  ADV=1,PART=1,ADP=2 & ADP=3 \\
 \bottomrule
 \end{tabular}
 }
 \caption{Each cell is the tag selected for German token `zu' at each iteration. Gold output tag distribution for `zu' is \text{\small{ADP=194, PART=103, ADV=5, PROPN=5, ADJ=1}}.
 }
  \label{tab:iteration}
  \vspace{-.5em}
 \end{table}

{
\algrenewcommand\algorithmicindent{1.2em}%
\begin{algorithm}[t]
    \caption{\textsc{Confusion-Reducing AL  
    }}
    \label{euclid}
    \begin{algorithmic}[1]
    \State $D \gets \textit{unlabeled set of sequences}$
    \State $Z \gets \textit{vocabulary}$
    \State $\mathcal{J} \gets \textit{output tag-set}$
    \State $b \gets \textit{active learning batch size}$
    \State $P_{\theta}(y_{i,t} = j \mid \mathbf{x_i}) \gets \textit{marginal probability}$
    \State $p_{i,t,j} :=  P_{\theta}(y_{i,t} = j \mid \mathbf{x_i})$
    \For {$\mathbf{x_i} \in D$}
    \For {($x_{i,t}) \in \mathbf{x_i}$}
    \State {$z \leftarrow x_{i,t}$}
    \State $S_{\textsc{cral}}(z)\!\leftarrow\!S_{\textsc{cral}}(z)\!+\!(1-p_{i,t,\hat{y}_{i,t}})$

    \State $\hat{j}\!\leftarrow\!\argmax_{j\in\!\mathcal{J}\setminus \{\hat{y}_{i,t}\}}p_{i,t,j}$
    \State $O_{\textsc{cral}}(z,\hat{j})\!\leftarrow\!O_{\textsc{cral}}(z,\hat{j})\!+1$
    \EndFor
    \EndFor
    \State $X_{\textsc{init}} \leftarrow {\rm b\text{-}}\argmax_{z \in Z} S_{\textsc{cral}}(z)$
    \For {$z_k \in X_{\textsc{init}}$}
    \State $j_k \leftarrow  \argmax_{j \in \mathcal{J}} O_{\textsc{cral}}(z_k,j) $ 
    \For {$x_{i,t} \in D \text{ s.t. } x_{i,t} = z_k $}
    \State $\mathbf{c}_{x_{i,t}} \leftarrow \text{enc}(x_{i,t})$
    \State $W_{x_{i,t}}\!= p_{i,t,j_k} * \mathbf{c}_{x_{i,t}} $
    \EndFor 
    \State $X_{\textsc{label}}(z_k) = \text{\textsc{centroid}} \{ W_{x_{i,t} = z_k} \}$
    \EndFor 
    \end{algorithmic}
\end{algorithm}
}
\section{Confusion-Reducing Active Learning}
\label{sec:al}
To address the limitations of the existing methods, we propose a \emph{confusion-reducing active learning} (\textsc{cral}) strategy, which aims at reducing the confusion between the output tags. In order to combine both ``informativeness'' and ``representativeness'', we follow a two-step algorithm:
\begin{enumerate}[leftmargin=*]
    \item \textbf{Find the most confusing word types.}
    The goal of this step is to find $b$ word types which would \emph{maximally reduce} the model confusion within the output tags. For each token $x_{i,t}$ in the unlabeled sequence $\mathbf{x_i} \in D$, we first define the confusion as the sum of probability $ P_{\theta}(y_{i,t}\!=\!j \mid \mathbf{x_i})$ of all output tags $\mathcal{J}$ other than the highest probability output tag $\hat{y}_{i,t}$:
    \begin{align*}
    S_{\textsc{conf}}(x_{i,t}) &=  1-P_{\theta}(y_{i,t}=\hat{y}_{i,t} \mid \mathbf{x_i}),
    \end{align*}
    then sum this over all instances of type $z$:
    \begin{align*}
    S_{\textsc{cral}}(z) &= \sum_{\mathbf{x_i} \in D} \sum_{x_{i,t}=z} S_{\textsc{conf}}(x_{i,t}).
    \end{align*}
    Again selecting the top $b$ types having the highest score (given by ${\rm b\text{-}}\argmax$) gives us the most confusing word types $(X_{\textsc{init}})$. For each token, we also store the output tag that had the second highest probability which we refer to as the ``most confusing output tag'' for a particular $x_{i,t}$:
  \[O(x_{i,t},j)\!=\!\begin{cases} 
      1 & \text{if } j\!=\!\argmax_{j\in\!\mathcal{J}\setminus \{\hat{y}_{i,t}\}}p_{i,t,j} \\
      0 & \text{otherwise.}
  \end{cases} \]
    
    For each word type $z$, we aggregate the frequency of the most confusing output tag across all token occurrences:
    \[ O_{\textsc{cral}}(z,j) = \sum_{\mathbf{x_i} \in D} \sum_{x_{i,t}=z} O(x_{i,t},j), \]
    and compute the output tag with the highest frequency as the most confusing output tag for type $z$: \[L(z) = \argmax_{j\!\in\!\mathcal{J}} O_{\textsc{cral}}(z,j). \]  For each of the top $b$ most confusing word types, we retrieve its most confusing output tag resulting in type-tag pairs given by $L_{\textsc{init}}=\{\langle z_1, j_1\rangle, \cdots \langle z_b, j_b\rangle \}$. This process is illustrated in steps 7--14 in Algorithm~\ref{euclid}.
    
    \item \textbf{Find the most representative token instances}. Now that we have the most confusing type-tag pairs $L_{\textsc{init}}$, our final step is selecting the most representative token instances for annotation.  For each type-tag tuple $\langle z_k, j_k \rangle\!\in\!L_{\textsc{init}}$, we first retrieve contextualized representations for all token occurrences ($x_{i,t} = z_k$) of the word-type $z_k$  from the encoder of the POS model. We express this in shorthand as $\mathbf{c}_{i,t} := \text{enc}(x_{i,t})$. Since the true labels are unknown, there is no certain way of knowing which tokens have the ``most confusing output tag'' as the true label. Therefore, each token representation $\mathbf{c}_{i,t}$ is weighted with the model confidence of the most confusing tag $j_k$ given by 
    \[ W_{x_{i,t}} =P_{\theta}(y_{i,t}\!=j_k\!\mid\mathbf{x_i}) * \mathbf{c}_{i,t} \] Finally, the token instance that is closest to the centroid of this weighted token set becomes the most representative instance for annotation. Going forward, we also refer to the most representative token instance  as the centroid for simplicity.\footnote{\citet{sener2017active} describe why choosing the centroid is a good approximation of representativeness. They pose AL  as a core-set selection problem where a core set is the subset of data on which the model if trained closely matches the performance of the model trained on the entire dataset. They show that finding the core set is equivalent to choosing $b$ center points such that the largest distance between a data point and its nearest center is minimized. We take inspiration from this result in using the centroid to be the most representative instance.}
    This process is repeated for each of the word-types $z_k$ resulting in  the to-label set $X_{\textsc{label}}$. This is illustrated in steps 14--19 in Algorithm~\ref{euclid}.
\end{enumerate}

\noindent During the annotation process, the selected representative tokens of each selected confusing word type are presented in context  similar to \citet{fang2017model, chaudhary-etal-2019-cmu}.

\section{Model and Training Regimen}
\label{sec:system}
Now that we have a method to select data for annotation, we present our POS tagger in Section \S \ref{sec:model}, followed by the training algorithm in Section \S \ref{sec:cvt}.

\subsection{Model Architecture}
\label{sec:model}
Our POS tagging model is a hierarchical neural conditional random field (CRF) tagger \cite{state, lample-etal-2016-neural, yang2017transfer} 
Each token $(\mathbf{x},t)$ from the input sequence $\mathbf{x}$ is first passed through a character-level Bi-LSTM, followed by a self-attention layer \cite{vaswani2017attention}, followed by another Bi-LSTM to capture information about subword structure of the words 
Finally, these character-level representations are fed into a token-level Bi-LSTM in order to create contextual representations $\mathbf{c_t}\!=\!\overrightarrow{\boldsymbol{h_t}}:\overleftarrow{\boldsymbol{h_t}}$, where $\overrightarrow{\boldsymbol{h_t}}$ and $\overleftarrow{\boldsymbol{h_t}}$ are the representations from the forward and backward LSTMs, and ``$:$'' denotes the concatenation operation. The encoded representations are then used by the CRF decoder to produce the output sequence.

Since we acquire token-level annotations, we cannot directly use the traditional CRF which expects a fully labeled sequence. Instead, we use a constrained CRF \cite{bellare2007learning} which computes the loss only for annotated tokens by marginalizing the un-annotated tokens, as has been used by prior token-level AL models \cite{fang2017model, chaudhary-etal-2019-cmu} as well. Given an input sequence $\mathbf{x}$ and a label sequence $\mathbf{y}$, traditional CRF computes the likelihood as follows:
\begin{align*}
    p_\theta(\mathbf{y}|\mathbf{x}) &= \frac{\prod_{t=1}^{N} \psi_t(y_{t-1},y_t,\mathbf{x},t)}{Z(\mathbf{x})},\\
    Z(\mathbf{x}) &= \sum_{\mathbf{y} \in \mathbf{Y}(N)} \prod_{t=1}^{N} \psi_t(y_{t-1},y_t,\mathbf{x},t),
\end{align*}
where $N$ is the length of the sequence, $\mathbf{Y}(N)$ denotes the set of all possible label sequences with length $N$. $\psi_t(y_{t-1},y_t, \mathbf{x})=\exp(\mathbf{W}^{T}_{y_{t-1},y_t}\mathbf{x}_t + \mathbf{b}_{y_{t-1},y_t})$ is the energy function where  $\mathbf{W}^{T}_{y_{t-1},y_t}$ and $\mathbf{b}_{y_{t-1},y_t}$ are the
weight vector and bias corresponding to label pair ($y_{t-1},y_t$) respectively. In constrained CRF training,  $\mathbf{Y_L}$ denotes the set of all possible sequences that are congruent with the observed annotations, and the likelihood is computed as:
$p_\theta(\mathbf{Y_L}|\mathbf{x}) = \sum_{\mathbf{y} \in \mathbf{Y_L}} p_\theta(\mathbf{y} | \mathbf{x})$.

\subsection{Cross-view Training Regimen}
\label{sec:cvt}
In order to further improve the above model, we apply cross-view training (CVT), a semi-supervised learning method~\cite{clark2018semi}. On unlabeled examples, CVT trains auxiliary prediction modules, which look at restricted ``views'' of an input sequence, to match the prediction from the full view. By forcing the auxiliary modules to match the full-view module, CVT improves the model's representation learning. Not only does it help in improving the downstream performance under low-resource conditions, but also improves the model calibration overall (\S \ref{sec:cvte}). Having a well-calibrated model is quite useful for AL, as a well-calibrated model tends to assign lower probabilities to ``true'' incorrect predictions which allows the AL measure to select these incorrect tokens for annotation.  

CVT is comprised of four auxiliary prediction modules, namely: the forward module $\theta_{fwd}$ which makes predictions without looking at the right of the current token, the backward module $\theta_{bwd}$ which makes predictions without looking at the left of the current token, the future module $\theta_{fut}$ which does not look at either the right context or the current token and, the past module $\theta_{pst}$ which does not look at either the left context or the current token. The token representations $\mathbf{c_t}$ for each module can be seen as follows:
\begin{align*}
    \mathbf{c_t^{fwd}}&=\overrightarrow{\boldsymbol{h_t}}, &\mathbf{c_t^{bwd}} &= \overleftarrow{\boldsymbol{h_t}}, &\mathbf{c_t^{full}} = \overrightarrow{\boldsymbol{h_t}}: \overleftarrow{\boldsymbol{h_t}}.\\
    \mathbf{c_t^{fut}}&=\overrightarrow{\boldsymbol{h}_{t-1}}, &\mathbf{c_t^{pst}} &= \overleftarrow{\boldsymbol{h}_{t+1}}.
\end{align*}
For an unlabeled sequence $\mathbf{x}$, the full-view model $\theta_{full}$ first produces soft targets $p_\theta({\mathbf{y}|\mathbf{x}})$ after inference. CVT matches the soft predictions from $V$ auxiliary modules by minimizing their KL-divergence.  Although CRF produces a probability distribution over all possible output sequences, for computational feasibility we compute the token-level KL-divergence using $p_\theta(y_t|\mathbf{x})$  which is the marginal probability distribution of token $(\mathbf{x},t)$ over all output tags $T$. This is calculated easily from the forward-backward algorithm: 
\begin{align*}
    \small{L_{\textsc{cvt}}} &\small{= \frac{1}{|D|} \sum_{\mathbf{x_i} \in D} \sum_{x_{i,t} \in \mathbf{x_i} } \sum_{v=1}^{V} KL (p_\theta^{full} || p_\theta^v)}\\
    p_\theta^{full}\!:\!&=\!P_\theta^{full}(y_{i,t}\!=\!j\!\mid\!\mathbf{x_i}) \text{ and } p_\theta^{v}\!:\!=\!P_{\theta}^v(y_{i,t}\!=\!j\!\mid\!\mathbf{x_i})
\end{align*}
 where $|D|$ is the total unlabeled examples in $D$. 
 
\subsection{Cross-Lingual Transfer Learning}
\label{sec:ctal}

Using the architecture described above, for any given target language we first train a POS model on a group of related high-resource languages. We then \textit{fine-tune} this pre-trained model on the newly acquired annotations on the target language, as obtained from an AL method. 
The objective of cross-lingual transfer learning is to warm-start the POS model on the target language. Several  methods have been proposed in the past including annotation projection \cite{zitouni2008mention}, model transfer using pre-trained models  such as m-BERT \cite{devlin2018bert}.  In this work our primary focus is on designing an active learning method, so we simply pre-train a POS model on a group of related high-resource languages \cite{cotterell-heigold-2017-cross} which is a computationally cheap solution, a crucial requirement for running multiple AL iterations.  Furthermore, recent work \cite{siddhant2019evaluating} has shown the advantage of pre-training using a selected set of related languages over a model pre-trained over all available languages.

Following this, for a given target language we first select a set of typologically related languages. An initial set of transfer languages is obtained using the  automated tool provided by \citet{lin2019choosing}, which leverages features such as  phylogenetic similarity,  typology, lexical overlap, and size of available data, in order to predict a list of optimal transfer languages. This list can be then refined using the experimenter's intuition. Finally, a POS model is trained on the concatenated corpora of the related languages. Similar to \citet{johnson2017google}, a language identification token is added at the beginning and end of each sequence.

\section{ Simulation Experiments}
\label{sec:setup}
In this section, we describe the simulation experiments used for evaluating our method. Under this setting, we use the provided training data as our unlabeled pool and simulate annotations by using the gold labels for each AL method.  
\paragraph{Datasets:} For the simulation experiments, we test on six typologically diverse languages: German, Swedish, North Sami, Persian, Ukrainian and Galician. We use data from the Universal Dependencies (UD) v2.3 \cite{nivre2016universal,nivre2018universal,kirov-etal-2018-unimorph} treebanks with the same train/dev/test split as proposed in \citet{mccarthy-etal-2018-marrying}. 
For each target language, the set of related languages used for pre-training is listed in Table~\ref{tab:dataset}. Persian and Urdu datasets being in the Perso-Arabic script, 
there is no orthography overlap along the transfer and the target languages. Therefore, for Persian we use uroman,\footnote{https://www.isi.edu/~ulf/uroman.html} a publicly available tool for romanization.
 \begin{table}[t]
 \centering
 \resizebox{\columnwidth}{!}{
 \begin{tabular}{@{}l|l@{}}
 \textbf{\textsc{target language }} & \textbf{\textsc{transfer languages (treebank)}}\\
 \midrule
 German (de-gsd) & English (en-ewt) + Dutch (nl-alpino)\\
 Swedish (sv-lines) & Norwegian (no-nynorsk) + Danish (da-ddt)\\
 North Sami (sme-giella) & Finnish (fi-ftb)\\
 Persian (fa-seraji) & Urdu (ur-udtb) + Russian (ru-gsd)\\
 Galician (gl-treegal) & Spanish (es-gsd) + Portuguese (pt-gsd)\\
 Ukrainian (uk-iu) & Russian (ru-gsd)\\
\midrule 
Griko & Greek (el-gdt) + Italian (it-postwita)\\
 \bottomrule
 \end{tabular}
 }
 \caption{Dataset details describing the group of related languages over which the model was pre-trained for a given target language. 
 }
   \label{tab:dataset}
 \vspace{-1em}
 \end{table}

\paragraph{Baselines:}
As described in Section \S \ref{sec:exwork}, we compare our proposed method (\textsc{cral}) with \emph{Uncertainty Sampling} ({\textsc{uns}}) and  \emph{Query-by-commitee} ({\textsc{qbc}}). We also compare with a random baseline ({\textsc{rand}}) that selects tokens randomly from the unlabeled data $D$. For \textsc{qbc}, we use the following committee of models $C = \{ \theta_{fwd}, \theta_{bwd}, \theta_{full} \}$, where $\theta_i$ are the  CVT views (\S \ref{sec:cvt}). We do not include the $\theta_{fut}$ and $\theta_{pst}$ as they are much weaker in comparison to the other views.\footnote{We chose CVT views for \textsc{qbc} over the ensemble for computational reasons. Training 3 models independently would require three times the computation. Given that for each language we run 20 experiments amounting to a total of 120 experiments, reducing the computational burden was preferred.}
For \textsc{cral}, \textsc{uns} and \textsc{rand}, we use the full model view.

\paragraph{Model Hyperparameters:}
We use a hidden size of 25 for the character Bi-LSTM, 100 for the modeling layer and 200 for the token-level Bi-LSTM. Character embeddings are 30-dimensional and are randomly initialized. We apply a dropout of 0.3 to the character embeddings before inputting to the Bi-LSTM. A further 0.5 dropout is applied to the output vectors of all Bi-LSTMs. The model is trained using the SGD optimizer with learning rate of 0.015. The model is trained till convergence over a validation set. 
\paragraph{Active Learning parameters:}
For all AL methods, we acquire annotations in batches of 50 and run 20 simulation experiments resulting in a total of 1000 tokens annotated for each method. We pre-train the model using the above parameters and after acquiring annotations, we fine-tune it with a learning rate proportional to the number of sentences in the labeled data $lr=2.5e^{-5}|X_{\textsc{label}}|$.

\subsection{ Results}
\label{sec:simulation}

\begin{figure}[t]
 \centering
\includegraphics[width=\columnwidth]{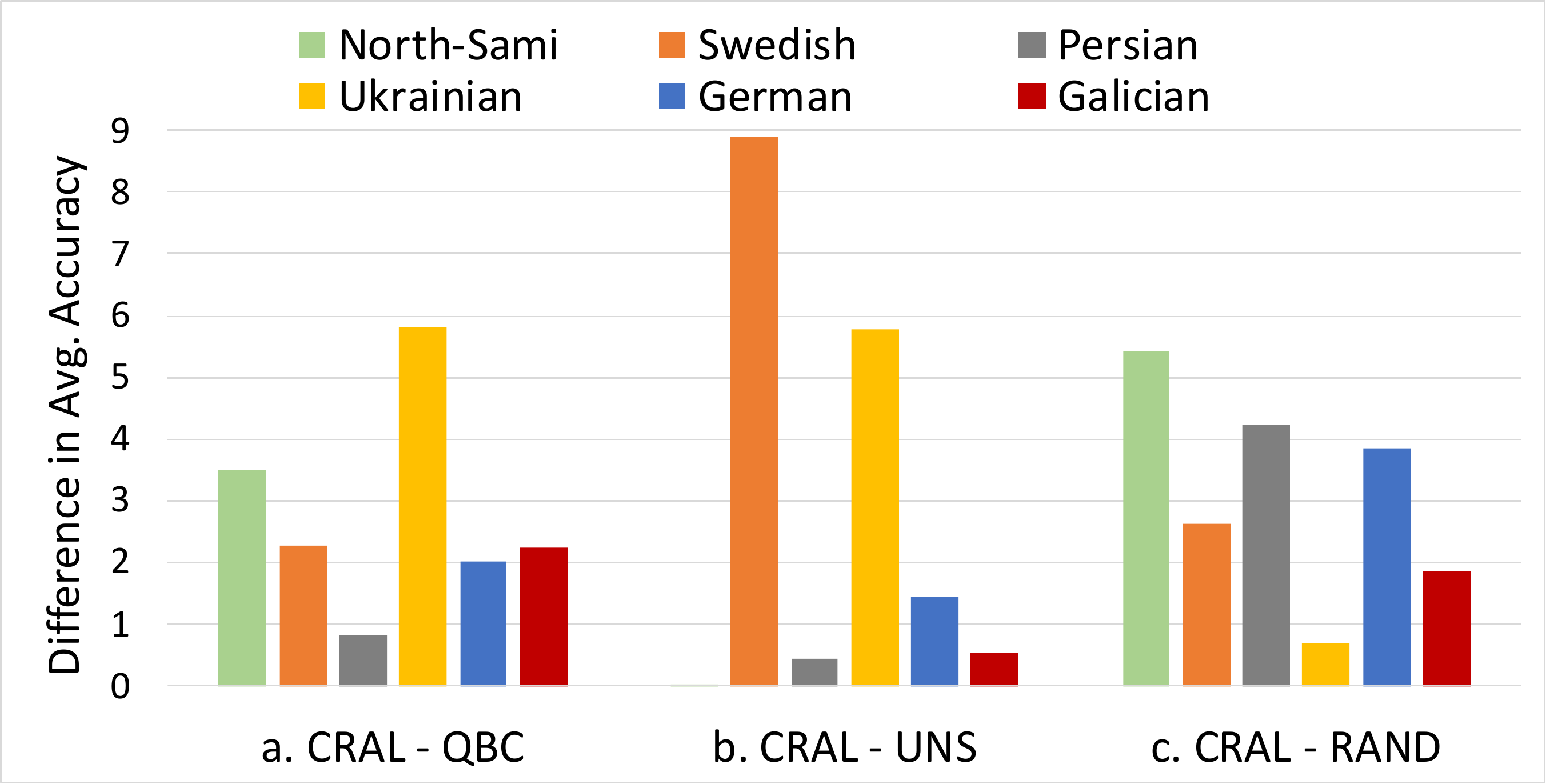}
\caption{\label{results} Our method (\textsc{cral}) outperforms existing AL methods for all six languages. y-axis is the difference in POS accuracy between \textsc{cral} and other AL methods, averaged across~20 iterations with batch size~50.}
\vspace{-1em}
\end{figure}

\begin{figure}[t]
 \centering
\includegraphics[width=\columnwidth]{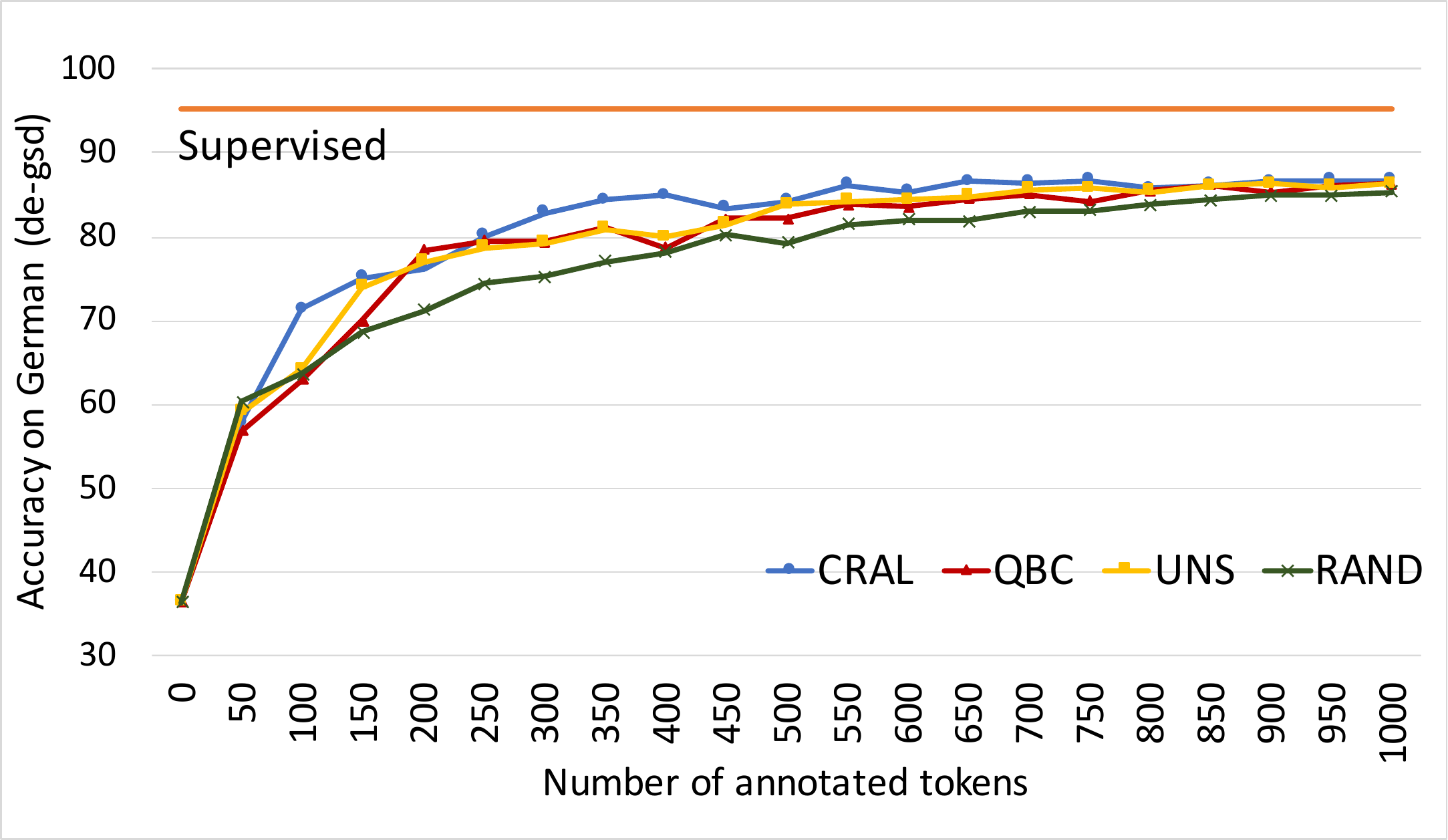}
 \caption{Comparison of the POS performance across the different methods for 20 AL iterations for German. }%
 \label{fig:ex3}%
\vspace{-1em}
\end{figure}

Figure~\ref{results} compares our proposed \textsc{cral} strategy with the existing baselines. Y-axis represents the difference in POS tagging performance between two AL methods and is measured by accuracy. The accuracy is averaged across~20 iterations. Across all six languages, our proposed method \textsc{cral} shows significant performance gains over the other methods. In Figure~\ref{fig:ex3} we plot the individual accuracy values across the 20 iterations for German and we see that our proposed method \textsc{cral} performs consistently better across multiple iterations. We also see that the zero-shot model on German (iteration-0) gets a decent warm start because of cross-lingual transfer from Dutch and English. 

\begin{figure}[t]
    \centering
    \includegraphics[width=0.5\textwidth]{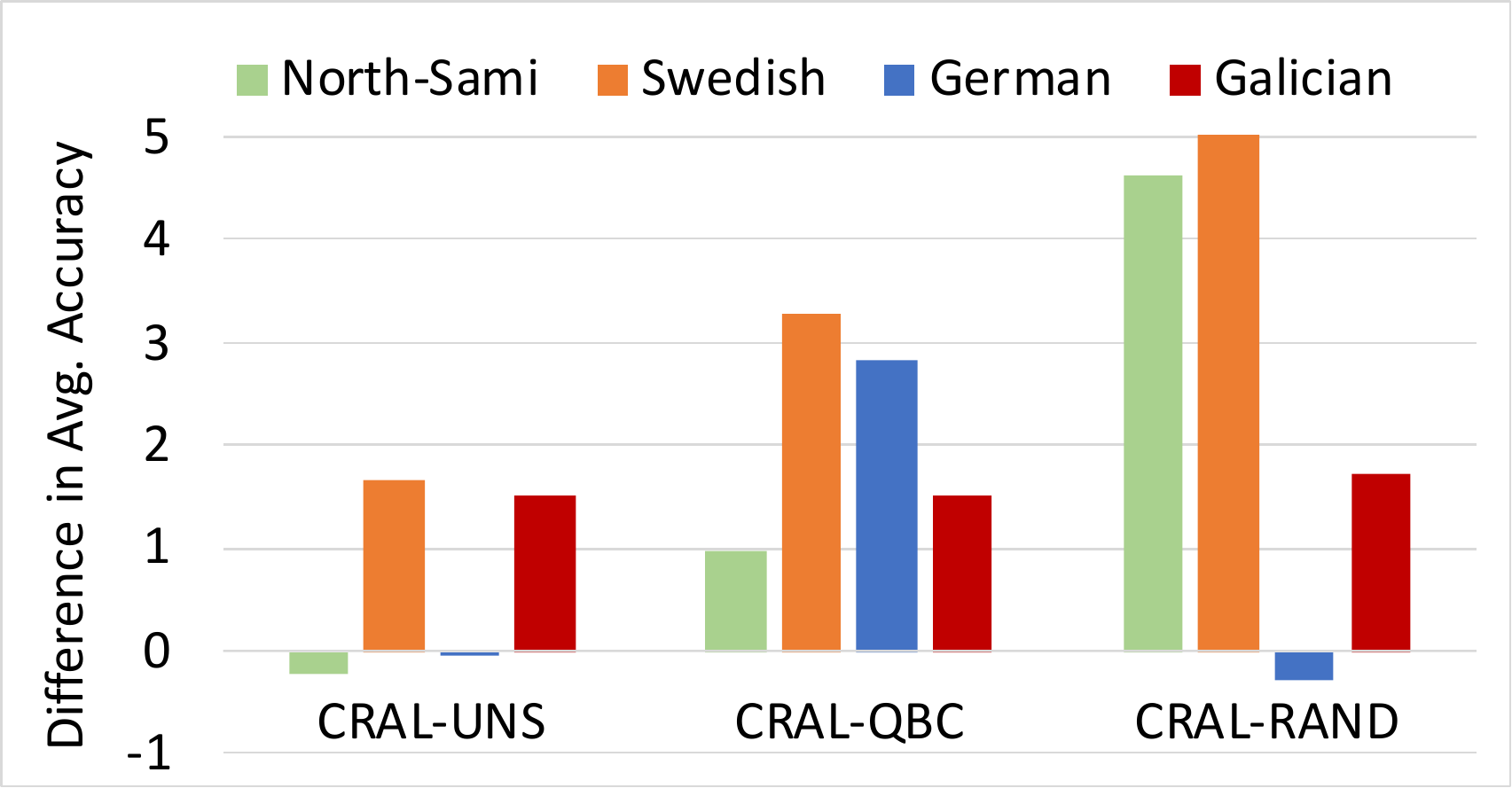}
    \caption{Comparing the difference in POS performance across the AL methods with BRNN/MLP architecture, averaged across 20 iterations.}
    \label{fig:simple_pos}
\end{figure}

Furthermore, to check how the performance of the AL methods is affected by the underlying POS tagger architecture, we conduct additional experiments with a different architecture. We replace the CRF layer with a linear layer and use token level softmax to predict the tags, keeping the encoder as before. We present the results for four (North Sami, Swedish, German, Galician) of the six languages in Figure~\ref{fig:simple_pos}.
Our proposed method \textsc{cral} still always outperforms \textsc{qbc}. We observe that only for North Sami, \textsc{uns} outperforms \textsc{cral}, which is similar to the results obtained from  BRNN/CRF architecture where the \textsc{cral} performs at par with \textsc{uns}.


\subsection{ Analysis}
\label{sec:qual}

\begin{table}[t]
\small
 \centering
 \resizebox{\columnwidth}{!}{
 \begin{tabular}{l|c|c|c}
 \textbf{\textsc{target language }}  & \textbf{\textsc{uns}} & \textbf{\textsc{qbc}}  & \textbf{\textsc{cral}} \\
 \midrule
 German  & 74 \% & 76 \% & 82\%  \\
 Swedish  & 56 \% & 54 \% & 62 \%\\
 North-Sami & 10 \% & 12 \% & 14   \%\\
 Persian &  50 \% & 46 \% & 46 \%  \\
 Galician  & 40 \% & 42 \% & 44 \%  \\
 Ukrainian & 38 \% & 48 \% & 48 \% \\
 \bottomrule
 \end{tabular}
 }
 \caption{Percentage of syncretic word types in the first iteration of active learning (consisting of 50 types). }
   \label{tab:qual}
 \vspace{-.5em}
 \end{table}
In the previous section,  we compared the different AL methods by measuring the average POS accuracy. In this section, we perform intrinsic evaluation to compare the quality of the selected data on two aspects:

\paragraph{How similar are the selected and the true data distributions? }

To measure this similarity, we compare the output tag distribution for each word type in the selected data with the tag distribution in the gold data. This evaluation is necessary because there are significant number of syncretic word types in the selected data as seen in Table \ref{tab:qual}. To recap,  \emph{syncretic} word types are word types that can have multiple POS tags based on context. 
We compute the Wasserstein distance (a metric to compute distance between two probability distributions) between the annotated tag distribution and the true tag distribution for each word type $z$.
\[ WD(z)= \sum_{j \in \mathcal{J}_z} p_j^{\textsc{AL}}(z) - p_j^{*}(z) \] where $\mathcal{J}_z$ is the set of output tags for a word type $z$ in the selected active learning data.
$p_j^{\textsc{AL}}(z)$ denotes the proportion of tokens  annotated with tag $j$ in the selected data and $p_j^{*}$ is the proportion of tokens having tag $j$ in the entire gold data. Lower Wasserstein distance suggests high similarity between the selected tag distribution and output tag distribution.
Given that each iteration selects unique tokens, this distance can be computed after each of $n$ iterations. 
Table \ref{tab:qual1} shows that our proposed strategy \textsc{cral} selects data which closely matches the gold data distribution for four out of the six languages.

\begin{table}[t]
\small
 \centering
 \resizebox{\columnwidth}{!}{
 \begin{tabular}{l|c|c|c}
 \textbf{\textsc{target language }} & \textbf{\textsc{cral}} & \textbf{\textsc{uns}} & \textbf{\textsc{qbc}} \\
 \toprule
 German & \textbf{0.0465} & 0.0801 & 0.0849 \\
 Swedish & \textbf{0.0811} & 0.1196 & 0.1013 \\
 North Sami & \textbf{0.0270} & 0.0328 & 0.0346 \\
 Persian & \textbf{0.0384} &  0.0583 & 0.0444  \\
 Galician & 0.0722 & 0.0953 & \textbf{0.0674} \\
 Ukrainian & 0.0770& 0.1067 & \textbf{0.0665} \\
 \bottomrule
 \end{tabular}
 }
 \caption{Wasserstein distance between the output tag distributions of the selected data and the gold data, lower the better. The above results are after 200 annotated tokens i.e. four AL iterations.  }
   \label{tab:qual1}
 \end{table} 
 
\begin{figure}[t]
\small
\includegraphics[width=\columnwidth]{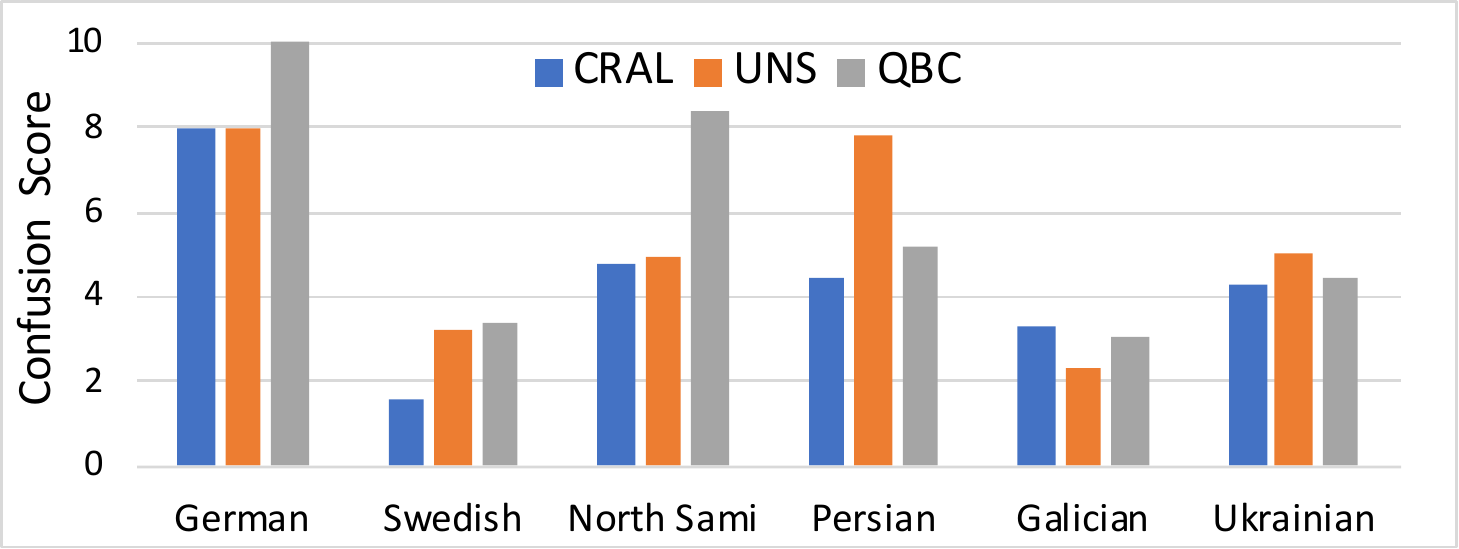}
\vspace{-1em}
\caption{\label{confu} Confusion score measures the percentage of correct predictions in the first iteration which were incorrectly predicted in the second iterations. Lower values suggest that the selected annotations in the subsequent iterations cause less damage on the model trained on the existing annotations.}
\vspace{-1em}
\end{figure}

\paragraph{How effective is the AL method in reducing confusion across iterations? } 
Across iterations, as more data is acquired we expect the incorrect predictions from the previous iterations to be rectified in the subsequent iterations, ideally without damaging the accuracy of existing predictions.  However, as seen in Table \ref{tab:qual}, the AL methods have a tendency to select syncretic word types which suggests that across multiple iterations the same word types could get selected albeit under a different context. This could lead to more confusion thereby damaging the existing accuracy if the selected type is not a good representative of its annotated tag.  Therefore, we calculate the number of existing correct predictions  which were incorrectly predicted in the subsequent iteration, and  present the results in Figure~\ref{confu}. A lower value suggests that the AL method was effective in improving overall accuracy without damaging the accuracy from existing annotations, and thereby was successful in reducing confusion. From Figure~\ref{confu}, the proposed strategy \textsc{cral} is clearly more effective than the others in most cases in reducing confusion across iterations.

 \begin{figure}[t]
\small
\centering
\includegraphics[width=\columnwidth]{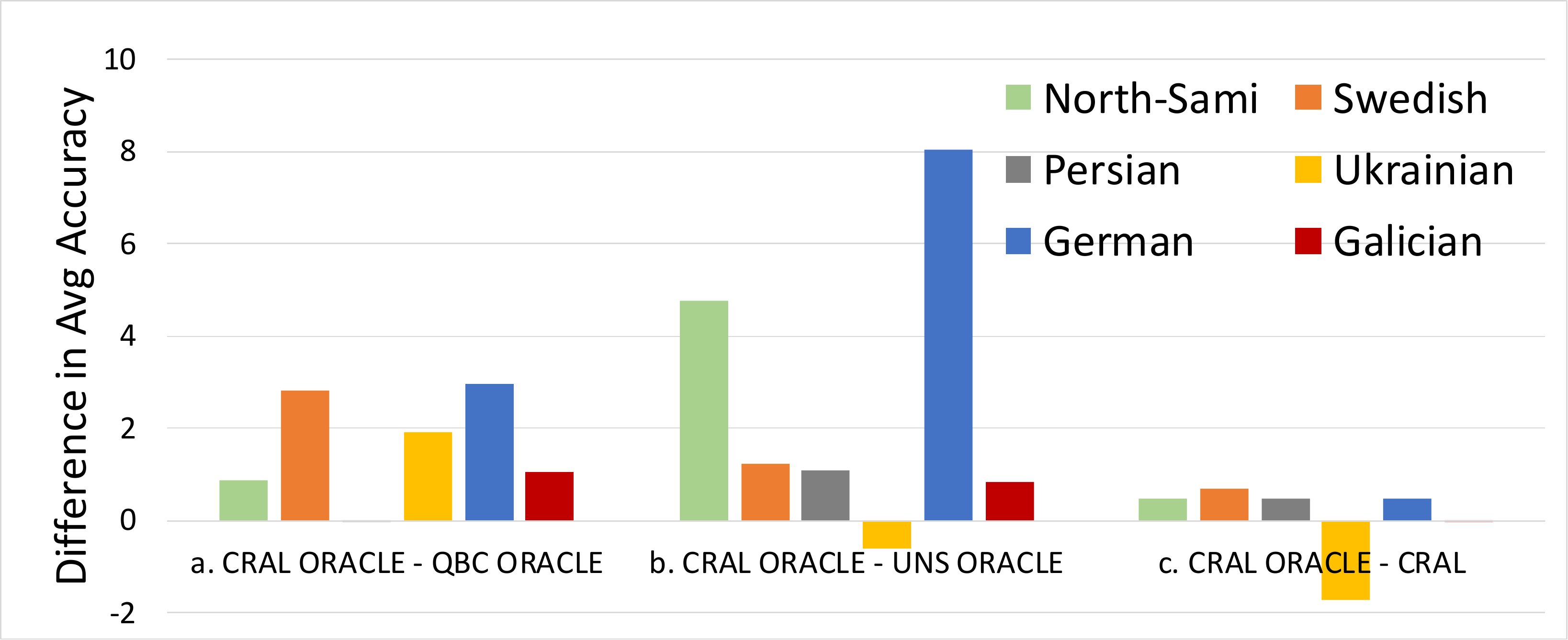}
\caption{\label{oracleresults} In the \textit{oracle} setting, our method (\textsc{cral-oracle}) outperforms \textsc{uns-oracle} and \textsc{qbc-oracle} in most cases, while the non-oracle \textsc{cral} matches the performance of its oracle counterpart. y-axis measures the difference in average accuracy across 20 iterations  between the methods being compared.} 
\vspace{-.5em}
\end{figure}

 \subsection{Oracle Results}
 \label{sec:oracler}
In order to check how close to optimal our proposed method \textsc{cral} is, we conduct ``oracle'' comparisons, where we have access to the gold labels during data selection. The oracle versions of existing methods \textsc{uns-oracle} and \textsc{qbc-oracle } are already described in Section \S \ref{sec:exwork}. For our proposed method \textsc{cral}, we construct the oracle version as follows:
    \paragraph{\textbf{\textsc{cral-oracle}}:}  Select the types having the highest incorrect predictions. Within each type, select that output tag which is most incorrectly predicted. This gives the most confusing output tag for a given word type. From the tokens having the most confusing output tag,  select the token representative by taking the centroid of their respective contextualized representations, similar to the procedure described in Section \S \ref{sec:al}.

Figure \ref{oracleresults} compares the performance gain of the POS model trained using \textsc{cral-oracle} over \textsc{uns-oracle} and \textsc{qbc-oracle}  (Figure \ref{oracleresults}.a, \ref{oracleresults}.b). 
Even under the ``oracle'' setting, our proposed method performs consistently better across all languages (except Ukrainian), unlike the existing methods as seen in Figure \ref{oracle}. \textsc{cral} closely matches the performance of its corresponding ``oracle'' \textsc{cral-oracle} (Figure \ref{oracleresults}.c) which suggests that the proposed method is close to an optimal AL method. However, we note that \textsc{cral-oracle} is not a ``true'' upper bound as for Ukrainian it does not out-perform \textsc{cral}. We find that for Ukrainian, up to 250 tokens, the oracle method outperforms the non-oracle method after which it under-performs. We hypothesize that this inconsistency is due to noisy annotations in Ukrainian. On analysis we found that the oracle method predicts numerals as NUM but in the gold data some of them are annotated as ADJ. We also find several tokens to have punctuations and numbers mixed with the letters.%
\footnote{This is also noted in the UD page: \url{https://universaldependencies.org/treebanks/uk_iu/index.html}}

\begin{figure}[t]
\small
\centering
\includegraphics[width=\columnwidth]{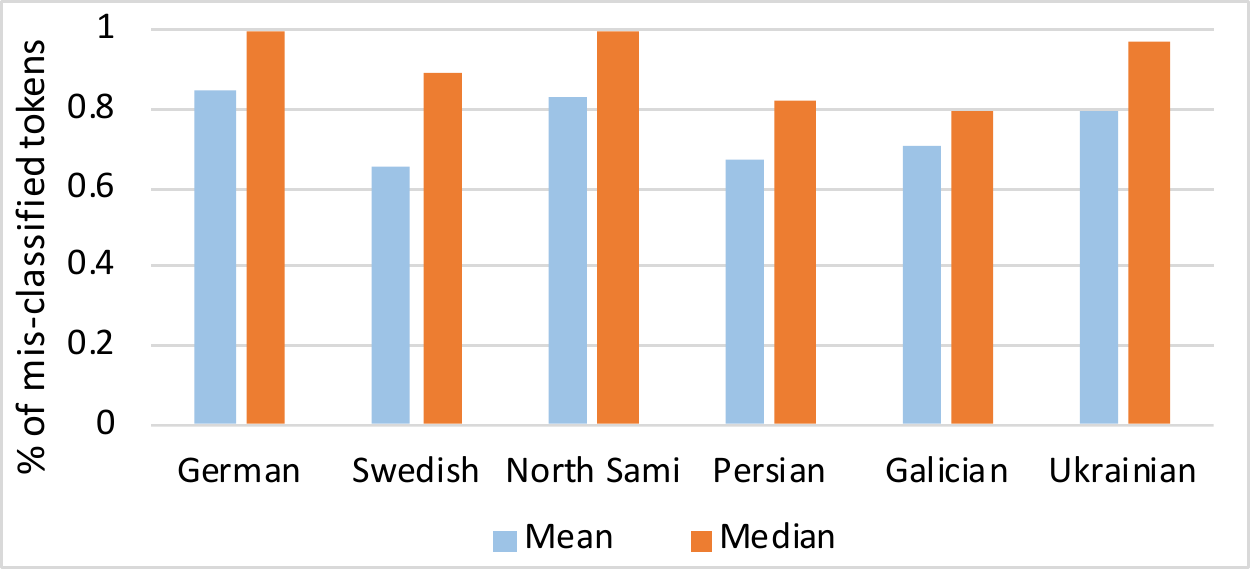}
\caption{\label{cralanaly} 
We report the mean and median of $\mathbf{p}$ over all the 50 token-tag pairs selected by the first AL iteration of \textsc{cral}. We see that across all languages majority of the token-tag pairs satisfy the criteria of using weighted representations with centroid for token selection.} 
\vspace{-1em}
\end{figure}

In order to verify whether \textsc{cral} is accurately selecting data at near-oracle levels, we analyze the intermediate steps leading to the data selection.
For each selected word type $z \in X_{\textsc{label}}$, we analyze how well our proposed method of weighting encoder representations with the model confidence of the most confused tag and taking the centroid actually succeeds at ``representative'' token selection.
If this is indeed the case, tokens in the vicinity  of the centroid should also have the same ``most confused tag'' as their predicted label and thereby be mis-classfied instances. To verify this hypothesis we compare how many of the 100 tokens closest to the centroid (in the representation space) ($X_{\textsc{nn}}(z)$) are truly mis-classified. This score is given by $p(z)$ for each selected word-type $z$:

\[ X_{\textsc{nn}}(z) = {\rm b\text{-}}\argmin_{x_{i,t}=z \in D} |\mathbf{c}_{i,t} - \mathbf{c}_{z}| \]
\[
p(z) = \frac{| \hat{y}_{i,t} \neq y^{*}_{i,t}|}{|X_{\textsc{nn}}(z)|} \]
where $b=100$. $\mathbf{c}_{z}$ is the contextualized representation of the representative instance for the word-type $z$ i.e the centroid and $\mathbf{c}_{i,t}$ is the contextualized representation of $z$'s token instance $x_{i,t}$. $y^{*}_{i,t}$ and $\hat{y}_{i,t}$ are the true and predicted labels of  $x_{i,t}$. We report the average and median of $\mathbf{p}$ across all the selected tokens of the first AL iteration in Figure \ref{cralanaly}.  We  see that for all languages the median is high (i.e. $>0.8$) which suggests that the majority of the token-tag pairs satisfy this criteria, thus supporting the step of weighting the token representations and choosing the centroid for annotation. 

We also compare the percent of token-tag overlap between the data selected from \textsc{cral} with its oracle counterpart: \textsc{cral-oracle}. For the first AL iteration, the proposed method \textsc{cral} has more than 50\% overlap with the oracle method for all languages, providing some evidence as to why \textsc{cral} is matching the oracle performance.

\subsection{Effect of Cross-View Training}
\label{sec:cvte}
As mentioned in Section \S \ref{sec:cvt}, we use cross-view training (CVT) to not only improve our model overall but also to have a well-calibrated model which can be important for active learning. A model is well-calibrated when a model's predicted probabilities over the outcomes reflects the true probabilities over these outcomes (\citet{nixon2019measuring}).  We use Static Calibration Error (SCE), a metric proposed by \citet{nixon2019measuring}  to measure the model calibration.  SCE bins the model predictions separately for each output tag probability and computes  the calibration error within each bin which is averaged across all the bins  to produce a single score. For each output tag, bins are created by sorting the predictions based on the output class probability. Hence, the first~10\% are placed in bin~1, the next~10\% in bin~2, and so on. We conduct two ablation experiments to measure the effect of CVT.   First,  we train a joint POS model on English and Norwegian datasets using all available training data, and evaluate on the English test set. Second, we use this pre-trained model and fine-tune on 200 randomly sampled German data and evaluate on German test data. We train models with and without CVT, denoted by +/- in Table~\ref{tab:cvteval}. We find that with CVT results both in higher accuracy as well as lower calibration error (SCE). This effect of CVT is much more pronounced in the second experiment, which presents a low-resource scenario and is common in an active learning framework.

\begin{table}[t]
\small
 \centering
 \resizebox{\columnwidth}{!}{
 \begin{tabular}{l|l|c|c}
 \textbf{\textsc{experiment setting }} & \textbf{\textsc{cvt}} & \textbf{\textsc{sce}} & \textbf{\textsc{accuracy}} \\
 \toprule
\multirow{2}{*}{EN + NO $\rightarrow$ EN} & - &  0.0190 & 95.53 \\
   & + &  \textbf{0.0174} & \textbf{95.58} \\
\midrule
\multirow{2}{*}{EN + NO + DE-200 $\rightarrow$ DE}   & - &  0.1658 & 69.90 \\
  & + &  \textbf{0.1391} & \textbf{74.61} \\
 \bottomrule
 \end{tabular}
 }
 \caption{Evaluating the effect of CVT across two experimental settings. EN: English, NO: Norwegian, DE-200: 200 German annotations. Left of `$\rightarrow$' are the pre-training languages and the right of `$\rightarrow$' is the language on which this model is evaluated.  Accuracy measures the POS model performance (higher is better) and SCE  measures the model calibration (lower is better).}
 \label{tab:cvteval}
 \vspace{-1em}
 \end{table}

\section{Human Annotation Experiment}
\label{sec:humanannotation}
In this section, we apply our proposed approach on Griko, an endangered language  spoken by around 20 thousand people in southern Italy, in the Grec\`{i}a Salentina area southeast of Lecce. The only available online Griko corpus, referred to as UoI~\cite{grikodatabase},\footnote{\url{http://griko.project.uoi.gr}} consists of 330 utterances by nine native speakers having POS annotations. Additionally, \citet{anastasopoulos2018part} collected, processed and released 114 stories, of which only the first 10 stories were annotated by experts and have gold-standard annotations.
 We conduct human annotation experiments on the remaining un-annotated stories in order to  compare the different active learning methods. 
\paragraph{Setup:} We use Modern Greek and Italian as the two related languages to train our initial POS model.\footnote{With Italian being the dominant language in the region, code switching phenomena appear in the Griko corpora.} To further improve the model, we fine-tune on the UoI corpus which consists of 360 labeled sentences. We evaluate the AL performance on the 10 gold-labelled stories from \citet{anastasopoulos2018part}, of which the first two stories, comprising of 143 labeled sentences, are used as the validation set and the remaining~800 labeled sentences form the test set. We use the unannotated stories as our unlabeled pool.
We compare \textsc{cral} with \textsc{uns} and \textsc{qbc}, conducting three AL iterations for each method, where each iteration selects roughly 50 tokens for annotation. The annotations are provided by two linguists, familiar with Modern Greek and somewhat familiar with Griko. To familiarize the linguists with the annotation interface, a practice session was conducted in Modern Greek. In the interface, tokens that need to be annotated are highlighted and presented with their surrounding context. The linguist then simply selects the appropriate POS tag for each highlighted token. Since we do not have gold annotations for these experiments, we also obtained annotations from a third linguist who is more familiar with Griko grammar.

\paragraph{Results:}
\begin{table*}[t]
\small
\centering
 \resizebox{!}{!}{
 \begin{tabular}{l|c|c|c|c|c|c|c}
 \textbf{} & \textbf{\textsc{AL}}& \textbf{\textsc{iteration-0}} & \textbf{\textsc{iteration-1}} & \textbf{\textsc{iteration-2}} & \textbf{\textsc{iteration-3}} & \textbf{IA Agr.} & \textbf{WD}\\
 \midrule
\multirow{3}{*}{Linguist-1}  & \textsc{cral} & 52.93 & \textbf{63.42} (10) & 	\textbf{69.07} (10) & 65.16 (16) & 0.58 & 0.281\\
            & \textsc{qbc} & 52.93 & 55.82 (15) & 62.03 (17) & \textbf{66.51}   (15) & 0.68 & 0.243\\
            & \textsc{uns} & 52.93 & 56.14 (15) & 	57.04 (15)  & 65.73 (11) & 0.58 & 0.379\\
 \midrule
\multirow{3}{*}{Linguist-2} & \textsc{cral} & 52.93 & \textbf{61.24} (15) &	\textbf{67.24} (20) & \textbf{67.05} (18)   & 0.70 & 0.346 \\
           & \textsc{qbc} & 52.93 &56.52 (20) & 65.96 (20) & 66.71 (17)  & 0.72 & 0.245\\
           & \textsc{uns} & 52.93 & 55.45 (17) & 	58.80 (17) & 65.73 (20) & 0.70 & 0.363\\
\midrule
\multirow{2}{*}{Linguist-3 }  & \textsc{cral} & 52.93 & \textbf{65.63} & 	\textbf{69.17} & \textbf{68.09} & - & 0.159\\
\multirow{2}{*}{(Expert)}  & \textsc{qbc} & 52.93 & 60.50 & 65.69 & 	56.20& - & 0.170\\
 & \textsc{uns} & 52.93 & 58.51 & 	64.26 & 65.93 &-&0.125\\
 \bottomrule
 \end{tabular}
 }
 \caption{Griko test set POS accuracy after each AL annotation iteration. Each iteration consists of 50 token-level annotations. The number in parentheses is the time in minutes required for annotation. The \textsc{IA Agr.} column reports the inter-annotator agreement against the expert linguist for the first iteration. \textsc{WD} is the Wasserstein distance between the selected tokens and the test distribution.}
   \label{tab:humaneval}
  \vspace{-1em}
 \end{table*}
 
Table \ref{tab:humaneval} presents the results on three iterations for each AL method, with our proposed method CRAL outperforming the other methods in most cases. We note that we found several frequent tokens (i.e 863/13740 tokens) in the supposedly gold-standard Griko test data to be inconsistently annotated. Specifically, the original annotations did not distinguish between coordinating (\texttt{CCONJ}) and subordinating conjunctions (\texttt{SCONJ}), unlike the UD schema. As a result, when converting the test data to the UD schema all conjunctions where tagged as subordinating ones. Our annotation tool, however, allowed for either \texttt{CCONJ} or \texttt{SCONJ} as tags and the annotators did make use of them. With the help of a senior Griko linguist (Linguist-3), we identified a few types of conjunctions that are always coordinating: variations of `and' (\texttt{ce} and \texttt{c'}), and of `or' (\texttt{e} or \texttt{i}). We fixed these annotations and used them in our experiments. 

\begin{figure}[t]
    \centering
    \includegraphics[width=.9\columnwidth]{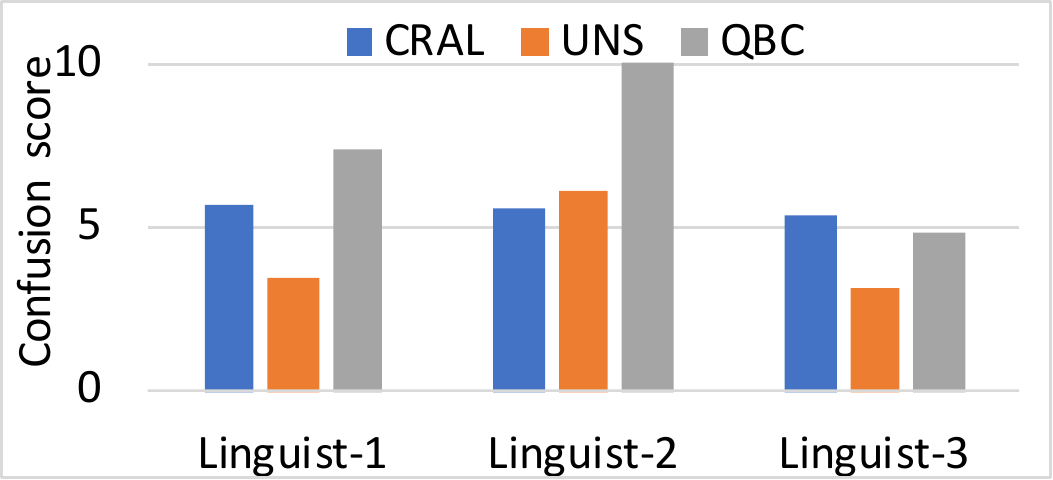}
    \caption{Confusion scores for the three Griko linguists. Lower values suggest that the selected annotations in the subsequent iterations cause less damage on the model trained on existing annotation. }
    \label{fig:griko_}
    \vspace{-1em}
\end{figure}

For Linguist-1, we observe a decrease in performance in Iteration-3. One possible reason for this decrease is attributed to Linguist-1's poor annotation quality which is also reflected in their low inter-annotator agreement scores. We observe a slight decrease for other linguists which we hypothesize is due to domain mismatch between the annotated data and the test data. In fact, the test set stories and the unlabeled ones originate from different time periods spanning a century, which can lead to slight differences in orthography and usage. For instance, after three  AL iterations, the token `i' had been annotated as CONJ twice and DET once, whereas in the test data all instances of `i' are annotated as DET. 
Similar to the simulation experiments, we compute the confusion score for all linguists in Figure \ref{fig:griko_}. We find that unlike in the simulation experiments, model trained with \textsc{uns} causes less damage on the existing annotations as compared to \textsc{cral}. However, we note that the model performance from the \textsc{uns} annotations is much lower than \textsc{cral} to begin with. 

We also compute the inter-annotator agreement at Iteration-1 with the expert (Linguist-3) (Table~\ref{tab:humaneval}).  We find that the agreement scores are lower than one would expect (c.f. the annotation test run on Modern Greek, for which we have gold annotations, yielded much higher inter-annotator agreement scores over 90\%). The justification probably lies with our annotators having limited knowledge of Griko grammar, while our AL methods require annotations for ambiguous and ``hard'' tokens. However, this is a common scenario in language documentation where often linguists are required to annotate in a language they are not very familiar with which makes this task even more challenging. We also recorded the annotation time needed by each linguist for each iteration in Table~\ref{tab:humaneval}. Compared to the \textsc{uns} method, the linguists annotated (avg.) 2.5 minutes faster using our proposed method which suggests that \textsc{uns} tends to select harder data instances for annotation.

Similar to the simulation experiments, we report the Wasserstein distance (WD)  for all linguists in Table 6. However, unlike in the simulation setting where the WD was computed with the gold training data, for the human experiments we do not have access to the gold annotations and therefore computed WD with the gold test data which however, is from a slightly different domain, which affects the results somewhat. We observe that \textsc{qbc} has lower WD scores for Linguist-1 and Linguist-2 and \textsc{uns} for Linguist-3. On further analysis, we find that even though \textsc{qbc} has lower WD, it also has the least coverage of the test data i.e. it has the fewest number of annotated tokens which are present in the test data as shown in Table \ref{tab:wd}. We would like to note that a lower WD score doesn't necessarily translate to better tagging accuracy because the WD metric is only informing us how good an AL strategy is in selecting data that matches closely the gold output tag distribution for that selected data. 

\begin{table}[t]
\small
\centering
 \begin{tabular}{l|l|l|l}
 \textbf{} & \textbf{\textsc{cral}} & \textbf{\textsc{uns}} & \textbf{\textsc{qbc}} \\
 \toprule
 Linguist-1 & 90 & 	95 & 	72 \\
Linguist-2 & 84	& 88 &	72 \\
Linguist-3 & 74	& 90 & 	61 \\
 \bottomrule
 \end{tabular}
 \caption{Each cell denotes the number of annotated tokens that are also present in the test data.}
  \label{tab:wd}

 \end{table}

\section{Related Work}
\label{sec:relatedwork}

\paragraph{Active Learning for POS tagging:} Active Learning (AL) has been widely-used for POS tagging. \cite{garrette-baldridge:2013:NAACL-HLT} use a graph-based label propagation to generalize initial POS annotations to the unlabeled corpus. Further, they find that under a constrained time setting, type-level annotations prove to be more useful than token-level annotations. In line with this, \cite{fang2017model} also select informative word types based on uncertainty sampling for low-resource POS tagging. They also construct a tag dictionary from these type-level annotations and then propagate the labels across the entire unlabeled corpus. However, in our initial analysis on uncertainty sampling, we found adding label-propagation harmed the accuracy in certain languages because of prevalent syncretism. \cite{ringger2007active} present different variations of uncertainty-sampling and query-by-committee methods for POS tagging. Similar to \cite{fang2017model}, they find uncertainty sampling with frequency bias to be the best strategy. \citet{settles2008analysis} present a nice survey on the different active learning strategies for sequence labeling tasks, whereas \citet{marcheggiani2014experimental} discuss the strategies for acquiring partially labeled data. \cite{sener2017active} propose a \emph{core-set} selection strategy aimed at finding the subset that is competitive across the unlabeled dataset. This work is most similar to ours with respect to using geometric center points as being the most representative.  However, to the best of our knowledge, none of the existing works are targeted at reducing confusion within the output classes. 

\paragraph{Low-resource POS tagging:}
Several cross-lingual transfer techniques have been used for improving low-resource POS tagging. \citet{cotterell-heigold-2017-cross,malaviya2018neural} train a joint neural model on related high-resource languages and find it be very effective on low-resource languages. The main advantage of these methods is that they do not require any parallel text or dictionaries. \citet{das2011unsupervised, tackstrom2013token,yarowsky2001inducing,nicolai-yarowsky-2019-learning} use annotation projection methods to project POS annotations from one language to another. 
However, annotation projection methods use parallel text, which often might not be of good quality for low-resource languages.

\section{Conclusion}
We have presented a novel active learning method for low-resource POS tagging which works by reducing confusion between output tags. Using simulation experiments across  six typologically diverse languages, we show that our confusion-reducing strategy achieves higher accuracy than existing methods. Further, we test our approach under a true setting of active learning where we ask linguists to document POS information for an endangered language, Griko. Despite being unfamiliar with the language, our proposed method achieves performance gains over the other methods in most iterations. For our next steps, we plan to explore the possibility of adapting our proposed method for \textit{complete} morphological analysis, which poses an even harder challenge for AL data selection due to the complexity of the task.

\section*{Acknowledgements}
The authors are grateful to the anonymous reviewers and the Action Editor who took the time to provide many interesting
comments that made the paper significantly better, and to Eleni Antonakaki and
Irini Amanaki, for participating in the human annotation experiments. This work is sponsored by
the Dr. Robert Sansom Fellowship, the Waibel Presidential Fellowship  and the 
National Science Foundation under grant 1761548.
\bibliography{tacl2018}
\bibliographystyle{acl_natbib}

\clearpage
\pagebreak
\clearpage

\end{document}